# Local Feature Descriptor Learning with Adaptive Siamese Network

Chong Huang[†], Qiong Liu[*], Yan-Ying Chen[*], Kwang-Ting (Tim) Cheng[†]

University of California, Santa Barbara[†], Fuji Xerox Palo Alto Laboratory[*]
chonghuang@umail.ucsb.edu, {Liu, yanying}@fxpal.com,
timcheng@ece.ucsb.edu

**Abstract.** Although the recent progress in deep neural network has led to the development of learnable local feature descriptors, there is no explicit answer for estimation of the necessary size of a neural network. Specifically, the local feature is represented in a low dimensional space, so the neural network should have more compact structure. The small networks required for local feature descriptor learning may be sensitive to initial conditions and learning parameters and more likely to become trapped in local minima. In order to address the above problem, we introduce an adaptive pruning Siamese Architecture based on neuron activation to learn local feature descriptors, making the network more computationally efficient with an improved recognition rate over more complex networks. Our experiments demonstrate that our learned local feature descriptors outperform the state-of-art methods in patch matching.

## 1 Introduction

The discussion about comparison between learned feature and traditional handcrafted feature never stops. The deep feature has achieved the superior performance for many classification tasks, even fine-grained object recognition. While the handcrafted local feature (e.g. SIFT) has been subject of study in computer vision for almost twenty years, the recent progress in deep neural network has led to a particular interest - learnable local feature descriptors.

While this direction has already proved to be quite fruitful, there are many challenges for local feature learning. One of the challenges is the estimation of the necessary size of a neural network. Specifically, the local feature is represented in a low dimensional space, so the neural network should have more compact structure. A common approach is to train successively smaller networks until the smallest one is found that will fit the data. This can be time consuming, in terms of training a number of networks. In addition, the smallest feasible networks may be sensitive to initial conditions and learning parameters and be more likely to become trapped in local minima. Although some work [1, 2, 3] has been used to estimate the necessary size of a system, they do not give implementation details about learning local feature. In other words, they do not answer how to choose a suitable network given a particular set of training data. Optimal Brain Damage [4] was proposed to reduce overfitting by deleting parameters with small "saliency", i.e. those whose deletion will have the least effect on the training error. After deletion, the network is retrained. This procedure can be iterated. Their work has given the information-theoretical proof that this strategy can achieve the nearly optimal network to fit a specific dataset.

In this work, we focus on the improvement of local feature descriptors based on Optimal Brain Damage. We propose a novel adaptive feature descriptor learning strategy as follows: Siamese architecture is utilized to minimize a discriminative loss function that drives the similarity metric to be small for matching pairs, and large for mismatching pairs. We start with a large configuration, remove the unimportant neurons iteratively until convergence, and finally output a compact network. We employ it to learn local feature descriptors in unsupervised scenarios.

The rest of the paper is organized as follows. In Section 2, we present our method. In Section 3, we discuss implementation details and our experimental results, respectively. We provide conclusions and future work in Section 4.

## 2  Method

We follow [5] to learn local feature in the Siamese Architecture, composed of two identical networks and one cost module. The input to the system is a pair of images and a label. The images are passed through the sub-networks, yielding two outputs which are passed to the cost module which produces the scalar energy. Contrastive loss function is utilized to train the network as follows:

$$E = \frac{1}{2N} \sum_{n=1}^{N} (y_n) d_n^2 + (1 - y_n) \max(1 - d_n, 0)^2$$

where $d_n$ is the square of distance between the $n$th pair of feature descriptors and $y_n$ is a binary label that indicates whether the $n$th pair matches or not.

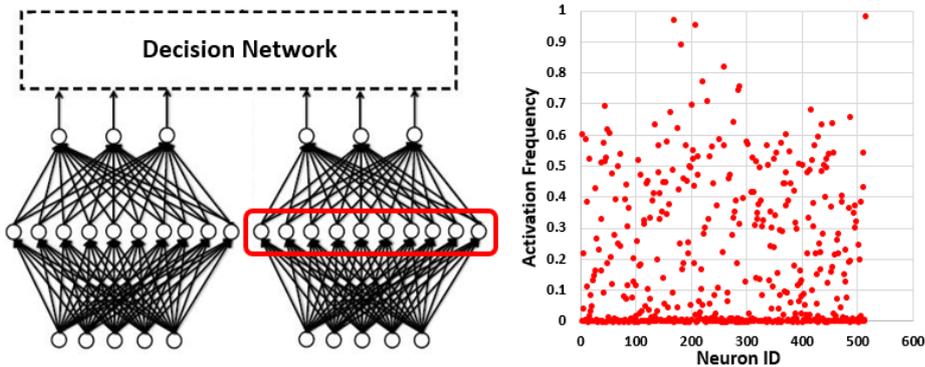

**Fig. 1.** (left) Siamese Architecture (right) The activation frequency of a fully-connected layer given 100k patches

However, the redundancy of neurons cannot be ignored. When we input one image patch for a neural network each time, we found that about 50% neurons are inactivated for each input. When we calculate the statistics for a fully-connected layer with 512 neurons given 100k image patches, the redundancy is more obvious. Fig.1 shows that 41% neurons' activation frequency is below 1%. The horizontal axis is the index of neurons and the vertical axis is the activation frequency. Although the sparsity is instrumental for the representative ability of feature, it is reasonable to remove the neurons with low sensitivity to earn more generation [4].

Based on the above observation, the intuition is to remove the redundant neurons from original network. That is, to train a network that is larger than necessary and then remove parts that are not needed. We choose the activation frequency of the neurons to decide which parts (i.e. neurons and connections) of network are not needed.

In summary, our algorithm can be carried out as follows:

Step 1: Choose a reasonable network architecture

Step 2: Train the network and update the weights

Step 3: Feed with validating pairs (i.e. 100k pairs in the experiment) to compute the activation frequency of all neurons.

Step 4: Remove the neurons whose activation frequency is below 1% and the related connection.

Step 5: Iterate to Step 2 until convergence.

## 3    Experiments

### 3.1    Dataset

We evaluated various methods using the UBC patch dataset [6]. UBC patch dataset includes three subsets (i.e. Liberty, Notre dame and Yosemite) with a total of more than 1.5 million patches of 64x64 pixels. We conduct the experiments on NVIDA Tesla K80 GPU.

### 3.2    Results

This dataset has provided the labeled image patches, so we focus on the adaptive pruning network on fully-connected layers. Because MatchNet [7] has achieved the best performance in this dataset, we start with their models [8] to prune the feature network.

**Table 1.** The layer parameters feature network before and after adaptive pruning

| Name | MatchNet | Proposed | Removed Ratio (%) |
|------|----------|----------|-------------------|
| FC1  | 4096     | 3874     | 6.4               |
| FC2  | 1024     | 646      | 37.1              |
| FC3  | 1024     | 698      | 31.9              |

Table 2. Error@95% (in percentage) of matching results over UBC dataset.

| Training | Notredame | Yosemite | Liberty | Yosemite | Liberty | Notredame | Mean |
|---|---|---|---|---|---|---|---|
| Test | Liberty | | Notredame | | Yosemite | | |
| MatchNet | 9.79 | 11.44 | 4.48 | 5.53 | 10.99 | 9.68 | 8.65 |
| Proposed | 9.17 | 10.09 | 4.46 | 5.51 | 10.80 | 9.30 | 8.22 |

We follow the evaluation protocol of MatchNet [7] and evaluate the patch-based matching before and after adaptive pruning matching. The evaluation metric is the false positive rate at 95% recall (Error@95%), the lower the better. Tab. 2 illustrates that the MatchNet can perform better after adaptive pruning. Although it seems that the improvement is subtle, Tab. 1 shows that the network becomes more compact.

## 4    Conclusion and Future Work

In this paper, we propose and evaluate a novel method for efficiently learning local feature descriptors in the Siamese network based on neuron activation. The experiments demonstrate that the learned descriptors can produce state-of-art results with more compact structure. In the future research, the pruning strategy on convolutional kernel will be explored further. This work will benefit more complex ANN-based applications on mobile device.